\documentclass[sigconf]{acmart}

\usepackage{balance}
\usepackage{multirow}
\usepackage{threeparttable}
\usepackage{amsmath,amssymb,amsfonts}

\def\BibTeX{{\rm B\kern-.05em{\sc i\kern-.025em b}\kern-.08emT\kern-.1667em\lower.7ex\hbox{E}\kern-.125emX}}
    
\copyrightyear{2019} 
\acmYear{2019}
\setcopyright{acmcopyright} 
\acmConference[MM '19]{Proceedings of the 27th ACM International Conference on Multimedia}{October 21--25, 2019}{Nice, France}
\acmBooktitle{Proceedings of the 27th ACM International Conference on Multimedia (MM '19), October 21--25, 2019, Nice, France}
\acmPrice{15.00}
\acmDOI{10.1145/3343031.3350975}
\acmISBN{978-1-4503-6889-6/19/10}

\begin{document}

\fancyhead{}

\title{Asynchronous Tracking-by-Detection on Adaptive Time Surfaces for Event-based Object Tracking}

\author{Haosheng Chen}
\authornote{With Fujian Key Laboratory of Sensing and Computing for Smart City, School of Informatics, Xiamen University, China.}
\affiliation{
	\institution{Xiamen University}
	\city{Xiamen}
	\country{China}
}
\email{haoshengchen@stu.xmu.edu.cn}
\author{Qiangqiang Wu}
\authornotemark[1]
\affiliation{
	\institution{Xiamen University}
	\city{Xiamen}
	\country{China}
}
\email{qiangwu@stu.xmu.edu.cn}
\author{Yanjie Liang}
\authornotemark[1]
\affiliation{
	\institution{Xiamen University}
	\city{Xiamen}
	\country{China}
}
\email{yanjieliang@yeah.net}
\author{Xinbo Gao}
\affiliation{
	\institution{Xidian University}
	\city{Xi'an}
	\country{China}
}
\email{xbgao@mail.xidian.edu.cn}
\author{Hanzi Wang}
\authornotemark[1]
\authornote{The corresponding author.}
\affiliation{
	\institution{Xiamen University}
	\city{Xiamen}
	\country{China}
}
\email{wang.hanzi@gmail.com}

%
\renewcommand{\shortauthors}{Chen, et al.}

%
\begin{abstract}
Event cameras, which are asynchronous bio-inspired vision sensors, have shown great potential in a variety of situations, such as fast motion and low illumination scenes. However, most of the event-based object tracking methods are designed for scenarios with untextured objects and uncluttered backgrounds. There are few event-based object tracking methods that support bounding box-based object tracking. The main idea behind this work is to propose an asynchronous Event-based Tracking-by-Detection (ETD) method for generic bounding box-based object tracking. To achieve this goal, we present an Adaptive Time-Surface with Linear Time Decay (ATSLTD) event-to-frame conversion algorithm, which asynchronously and effectively warps the spatio-temporal information of asynchronous retinal events to a sequence of ATSLTD frames with clear object contours. We feed the sequence of ATSLTD frames to the proposed ETD method to perform accurate and efficient object tracking, which leverages the high temporal resolution property of event cameras. We compare the proposed ETD method with seven popular object tracking methods, that are based on conventional cameras or event cameras, and two variants of ETD. The experimental results show the superiority of the proposed ETD method in handling various challenging environments.
\end{abstract}

%
%
%


%

%

%
\maketitle

\begin{figure}[t]
	\begin{center}
		\includegraphics[width=0.735\linewidth]{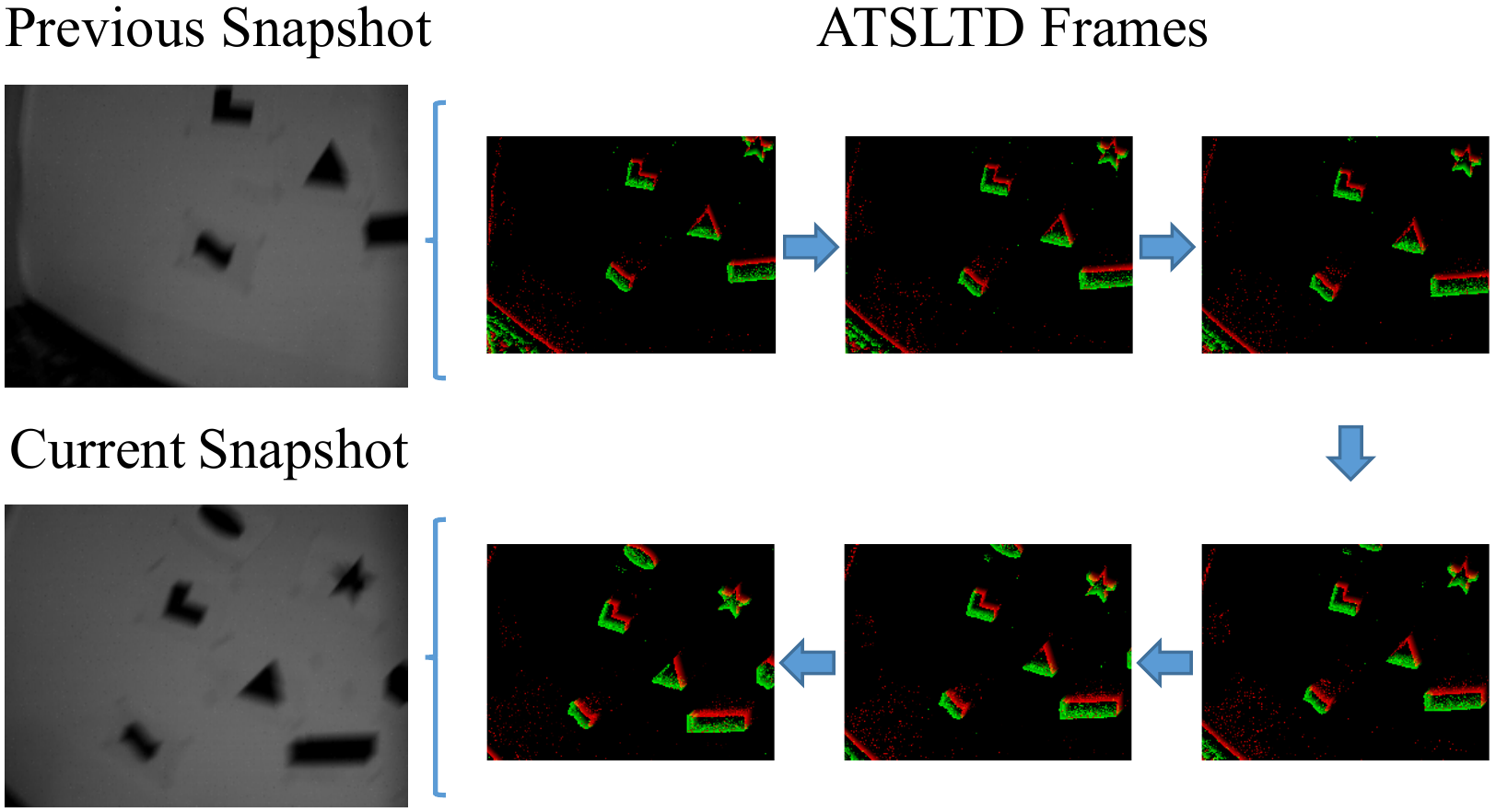}
	\end{center}
	\vspace{-3ex}
	\caption{An illustration of a sequence of ATSLTD frames. The left two images show two snapshots, while the right six images show the sequence of ATSLTD frames between the two snapshots.}
	\vspace{-3ex}
	\label{fig:retinalEvents}
\end{figure}


\section{Introduction}
Event cameras (e.g., DVS \cite{lichtsteiner2008128}, DAVIS \cite{brandli2014240} and ATIS \cite{posch2011qvga}) are bio-inspired silicon-retina visual sensors with very high dynamic range and temporal resolution ($>$120 dB, $<$1 ms). Thanks to the asynchronous nature of the biological retina, it can precisely and efficiently capture motion information \cite{oyster1968analysis,murphy2018old}, especially for motions caused by moving objects \cite{olveczky2003segregation,wild2018does}, in natural scenes. Inspired by the biological retina, unlike traditional cameras, asynchronous event cameras contain an array of independent pixels that asynchronously respond to pixel intensity changes in the environment. The asynchronous property of event cameras makes themselves suitable for object tracking and motion estimation.

If the intensity of a pixel on an event camera exponentially increases, the event camera will record an ``$On$'' event, which includes the coordinate of the event pixel and the current timestamp. In contrast, if the intensity of the pixel exponentially decreases, the event camera will record an ``$Off$'' event with the pixel coordinate and the current timestamp. Since the intensity changes are usually caused by object motions, event cameras can filter out non-motion information (for example, the static part of a scene) from their visual input, under stable illumination conditions or infrequent light variations. Thus event-based methods can acquire a clear and accurate clue about where object movement occurs, and save a lot of computational cost for searching moving objects.

Event cameras have achieved many successes on camera motion estimation and optical flow estimation (e.g., \cite{kim2016real,gallego2017event,zhu2018ev}). These successes have shown the superiority of event cameras on motion related tasks. However, there are only a few studies devoted to event-based object tracking, and most of these studies are designed for some special scenarios (e.g., \cite{pikatkowska2012spatiotemporal} is for the pedestrian tracking scenario). Therefore, in this paper, we study generic object tracking methods based on event cameras.

In this study, we present an asynchronous event-to-frame conversion algorithm, which asynchronously warps the spatio-temporal information in retinal events to a novel Adaptive Time-Surface with Linear Time Decay (ATSLTD) frame representation, as shown in Fig. \ref{fig:retinalEvents}. The conversion is driven by object motions: (1) Fast object motions will create more ATSLTD frames than slow object motions, which creates a relatively large spatio-temporal space for object tracking; (2) If the target object does not have enough large motion (displacement), there is no any generated ATSLTD frame, which increases the computational speed. Since the intensity changes and retinal events usually occur at the edges of the target object, the ATSLTD frames record clear and sharp object contours. Based on the generated ATSLTD frames, we propose an effective and efficient Event-based Tracking-by-Detection (ETD) method for event-based object tracking. 
The proposed ETD method can leverage the spatio-temporal consistency between adjacent ATSLTD frames and perform accurate and high-speed event-based object tracking. Overall, this study makes the following contributions:

\begin{itemize}
	\item We present an ATSLTD event-to-frame conversion algorithm to asynchronously warp the spatio-temporal information in retinal events, created by event cameras, to ATSLTD frames. The conversion is driven by object motions, and it does not rely on empirical cut-off thresholds.
	\item We introduce a new metric, named Non-Zero Grid Entropy (NZGE), to measure the amount of information in ATSLTD frames for adaptive event-to-frame conversion. And we calculate a confidence interval of NZGE values to perform the event-to-frame conversion algorithm asynchronously.
	\item We propose an Event-based Tracking-by-Detection (ETD) method to effectively and efficiently perform bounding box-based object tracking on the ATSLTD frames.
\end{itemize}

We evaluate the proposed ETD method on a mixed challenging event dataset consisting of a part of the event camera dataset and the extreme event dataset. The experimental results demonstrate the superiority of our ETD method when it is compared with several popular object tracking methods.

The rest of the paper is organized as follows. Section \ref{sec:RelatedWork} reviews the related work. Section \ref{sec:Methods} describes the ATSLTD event-to-frame conversion algorithm and the proposed ETD method. The performance of the ETD method is extensively evaluated and analyzed in Section \ref{sec:Experiments}, and conclusions are drawn in Section \ref{sec:Conclusion}.


\section{Related Work}
\label{sec:RelatedWork}
In this section, we firstly review some event-to-frame conversion works, then we introduce previous works on event-based motion estimation and object tracking.

\textbf{Event-to-frame conversion works.} There are several state-of-the-art works \cite{lagorce2017hots, liu2018adaptive, zhu2018ev,sironi2018hats, gallego2018unifying, maqueda2018event} proposed to convert asynchronous retinal events to synchronous frames. However, all these works rely on one or more empirical cut-off thresholds, which limits their applications. Among these works, the Time Surface-based algorithm \cite{lagorce2017hots} uses a fixed radius to compute the spatial neighborhood as its cut-off threshold. Because the spatial neighborhood is defined in the context of a timestamp map, the fixed radius can be treated as another form of the time window. The representations in \cite{zhu2018ev,sironi2018hats,gallego2018unifying,maqueda2018event} respectively use a fixed size of time window as their cut-off thresholds. The Adaptive Time-Slice representation in \cite{liu2018adaptive} uses either a constant event number or a fixed search radius as its cut-off threshold.

Using either a fixed time window or a constant event number as the cut-off threshold of sequential retinal events is an easy and straightforward way to perform an event-to-frame conversion. However, both types of cut-off thresholds (i.e., a fixed time window and a constant event number) cannot adapt to all circumstances. The generated event-based frames will be affected by the speed of motions and the texture of objects. In this study, we propose an adaptive and asynchronous event-to-frame conversion algorithm without using empirical cut-off thresholds.

\textbf{Event-based motion estimation and object tracking works.} Event-based works have achieved great success on motion related tasks. In the field of event-based optical flow estimation, \cite{bardow2016simultaneous} proposes a sliding window variational optimization algorithm to estimate the optical flow. \cite{gallego2018unifying} exploits the best point trajectories of the event data for optical flow estimation. The optical flow can also be estimated by a self-supervised deep neural network \cite{zhu2018ev}, or by a time-slice block-matching method \cite{liu2018adaptive}. In the field of event-based camera motion estimation, the 3D 6-DoF camera motion can be estimated by using an interleaved probabilistic filter \cite{kim2016real}, a photometric depth map \cite{gallego2017event}, or an image feature-based extended Kalman filter \cite{zihao2017event}. There also are some works in relation to event-based feature tracking \cite{gehrig2018asynchronous} and 3D reconstruction \cite{rebecq2018emvs}. From these motion related works, we can observe that event-based methods usually outperform conventional methods, especially when coping with fast motion and High Dynamic Range (HDR) scenes, due to the high temporal resolution and HDR properties of event cameras.

\begin{figure*}
	\begin{center}
		\includegraphics[width=0.76\linewidth]{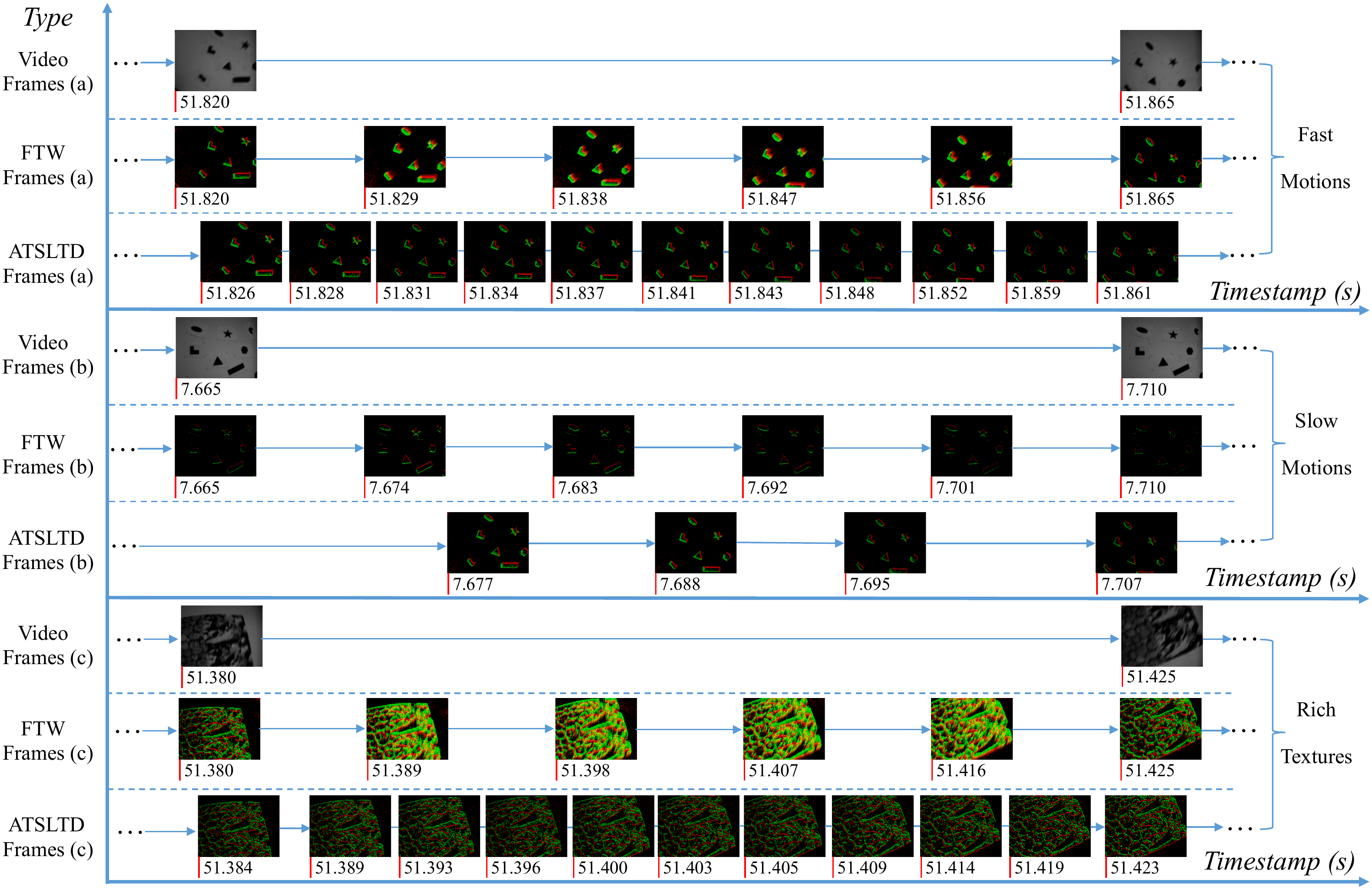}
	\end{center}
	\vspace{-3ex}
	\caption{Comparison of the Fixed Time Window (FTW) event-to-frame conversion algorithm and the proposed ATSLTD event-to-frame conversion algorithm. Video Frames (a-c) respectively show two snapshots, which are seen by a conventional camera before and after the events-to-frame conversion is implemented. FTW Frames (a-c) respectively show the generated FTW frames by the FTW event-to-frame conversion algorithm for fast object motion (a), slow object motion (b) and rich object texture (c) cases. ATSLTD Frames (a-c) respectively show the ATSLTD frames generated by the proposed ATSLTD event-to-frame conversion algorithm for the same cases. Red and green contours are respectively generated by ``$On$'' and ``$Off$'' events.}
	\label{fig:conversion}
	\vspace{-3ex}
\end{figure*}

However, compared with the conventional object tracking methods (such as \cite{bertinetto2016fully, danelljan2017eco, yang2017deep, zhang2017robust, gao2018osmo, liang2018robust}), there are only a few works devoted to event-based object tracking. The works can be roughly divided into two categories: i.e., clustering based works and non-clustering based works. The works in the first category require a clustering process to group retinal events into clusters. For example, \cite{litzenberger2006embedded} proposes a cluster-based object tracking method, which is inspired by the mean-shift algorithm \cite{comaniciu2000mean,comaniciu2002mean}. \cite{pikatkowska2012spatiotemporal} uses the Gaussian mixture model to group retinal events into clusters and track the clusters with occlusions. Similarly, \cite{camunas2017event} also relies on a clustering algorithm, which is a variant of \cite{litzenberger2006embedded}, to track objects with occlusions. \cite{glover2017robust} proposes a variant of the particle filters to track the clusters grouped by an event-based Hough transform algorithm \cite{ballard1981generalizing}. 

In the second category of works, \cite{Liu2016combined} proposes a particle filter-based object tracking method, which exploits the information from both video frames and retinal events. For some recent works, \cite{mitrokhin2018event} proposes a Kalman filter-based object tracking method, which uses a motion compensation algorithm. \cite{ramesh2018long} proposes a sliding window-based method for long-term object tracking. \cite{barranco2018real} presents a Kalman filter-based object tracking method for multi-target tracking. 

For the above-mentioned object tracking methods, we observe that most of them are designed at pixel level, which cannot support generic bounding box-based object tracking. In this study, we propose an event-based tracking-by-detection method to support generic bounding box-based object tracking.

\section{The Proposed Method}
\label{sec:Methods}
In this section, we present the Adaptive Time-Surface with Linear Time Decay (ATSLTD) event-to-frame conversion algorithm for the sequential retinal events generated by an event camera in Sec. \ref{subsec:ATSLTD}. Then we introduce a metric named Non-Zero Grid Entropy (NZGE), to measure the amount of information in generated ATSLTD frames and we compute a confidence interval of the NZGE values to perform the asynchronous ATSLTD event-to-frame conversion in Sec. \ref{subsec:ConfidenceInterval}. Finally, we propose an event-based tracking-by-detection (ETD) method which works on the ATSLTD frames in Sec. \ref{subsec:ETD}. Along with the ETD method, we also propose an event-based tracking recovery strategy for handling the tracking failure situation. Next, we will introduce the proposed method in detail.

\subsection{Adaptive Time-Surface with Linear Time Decay}
\label{subsec:ATSLTD}
Along with the pixel-level intensity changes that are caused by camera motions and object motions, sequential retinal events $\mathcal{E}$ are generated asynchronously by an event camera. Each event $e$ of $\mathcal{E}$ can be represented as a quadruple:
\begin{equation}
e \doteq (u, v, p, t),
\end{equation}
where $u$ and $v$ are the horizontal and vertical coordinates of $e$; $p$ indicates the polarity (\emph{On} or \emph{Off}) of $e$, and $t$ is the timestamp of $e$. Event cameras have very high spatio-temporal resolution and high HDR, therefore event-based object tracking methods can benefit a lot from the retinal event input captured by event cameras. Since each pixel of an event camera can occur independently in response to log intensity changes, the asynchronous nature of retinal events makes it difficult for conventional frame-based object detection and tracking methods to process the retinal events directly. As a result, we need an event-to-frame conversion algorithm, which warps the retinal events to an event-based frame representation.

Effective event-based frame representations for object detection and tracking should have sharp and clear object contours for moving objects. Meanwhile, the object contours should not have too much displacement, and should not be too sparse. The state-of-the-art event-to-frame conversion algorithms \cite{lagorce2017hots, liu2018adaptive, zhu2018ev,sironi2018hats, gallego2018unifying, maqueda2018event} use either a fixed time window or a constant event number as the cut-off threshold to perform the event-to-frame conversion. For the Fixed-Time-Window (FTW) event-to-frame conversion algorithms, we show some results obtained by an example FTW event-to-frame conversion algorithm using a fixed time window of 9 $ms$ in the FTW Frames (a-c) of Fig. \ref{fig:conversion}. The example conversion can create clear object contours for objects without highly speedy motions and complicated textures. However, the example algorithm generates sparse object contours for objects with slow motions (as shown in the FTW Frames (b) of Fig. \ref{fig:conversion}), and it generates blurred object contours with large contour displacement for the fast object motion and rich object texture situations (as shown in the FTW Frames (a) and (c) of Fig. \ref{fig:conversion}, respectively). From Fig. \ref{fig:conversion}, we can conclude that different object textures and speeds will result in different amounts of retinal events. Therefore, for the constant-event-number event-to-frame conversion algorithms, they are difficult to determine the constant event number. In addition, both types of the event-to-frame conversion algorithms are synchronous, which is incompatible with the asynchronous nature of event cameras.

In this work, we propose an event-to-frame conversion algorithm, which warps sequential retinal events to the Adaptive Time-Surface with Linear Time Decay (ATSLTD) frame representation. The ATSLTD representation also uses a time map to represent the spatio-temporal information in retinal events, which is similar to the Time-Surface representation in \cite{lagorce2017hots}. However, the main difference is that the proposed ATSLTD event-to-frame conversion algorithm is asynchronous. It is driven by object motions (faster object motions will create more ATSLTD frames than slower object motions), and it does not need any empirical cut-off thresholds for converting sequential retinal events to event-based frames. Moreover, we introduce an efficient linear time decay kernel to the proposed ATSLTD event-to-frame conversion algorithm for recording object contours on the ATSLTD frames.

When we implement the conversion of retinal events to the ATSLTD representation, the $i$-th ATSLTD frame $\mathcal{F}_{i}$ is initialized to a three-dimensional zero matrix $\mathcal{F}_{i} \in {\mathbb{N}^{h \times w \times 2}}$ at timestamp $\mathcal{T}_{i-1}$ when the previous ($i$-1)-th ATSLTD frame $\mathcal{F}_{i-1}$ is finished. Here $h$ and $w$ are the height and width of the event camera resolution, and the third dimension represents binary polarities of events (i.e., \emph{On} retinal events and \emph{Off} retinal events will be stored separately to avoid the mutual interference). Then each of the incoming retinal events will trigger an update on $\mathcal{F}_{i}$, which will multiply a linear time decay factor ${t_{k-1}}/{t_{{k}}}$ with $\mathcal{F}_{i}$ and then set a value of 255 to an element of $\mathcal{F}_{i}$ at the coordinates ${(u, v)}$ that correspond to the triggering event. For the $k$-th update caused by the $k$-th event ${e_{k} = \{u_k, v_k, p_k, t_k\}}$, the update is calculated as follow:
\begin{equation}
\label{fig:lineardecay}
\mathcal{F}_{i} = round(\mathcal{F}_{i}*{t_{k-1}}/{t_k}), \quad \mathcal{F}_{i}(u_k, v_k) = 255,
\end{equation}
where ${t_{k-1}}$ is the timestamp of the ($k$-1)-th event ${e_{k-1}}$ and ${t_{k}}$ is the timestamp of ${e_{k}}$. Finally, when the amount of information in $\mathcal{F}_{i}$ reaches the calculated confidence interval at the current timestamp $\mathcal{T}_{i}$, the ATSLTD event-to-frame conversion for $\mathcal{F}_{i}$ is finished and $\mathcal{F}_{i}$ becomes a new ATSLTD frame for object tracking.

Since pixel intensity changes usually occur at the edges of moving objects, retinal events will occur at the edges of moving objects in response to the pixel intensity changes. As a result, after the warping process, the linear time decay factor ${t_{k-1}}/{t_{{k}}}$ in Eq. (\ref{fig:lineardecay}) will form a pixel intensity decay at the edges of each moving object. The pixel intensity decay records a sequence of adjacent object contours (i.e., an object contour displacement) for the corresponding moving object, as shown in the ATSLTD Frames (a-c) of Fig. \ref{fig:conversion}. And a newly generated object contour of the sequential object contours has a higher pixel intensity than a previously generated object contour, which makes object contour-based detectors (e.g., EdgeBoxes \cite{zitnick2014edge}) pay more attention to the newly generated object contours in object detection. Therefore, the ATSLTD frame representation facilitates the proposed event-based tracking-by-detection method to detect and locate moving objects effectively and efficiently.

\begin{figure}[t]
	\begin{center}
		\includegraphics[width=0.85\linewidth]{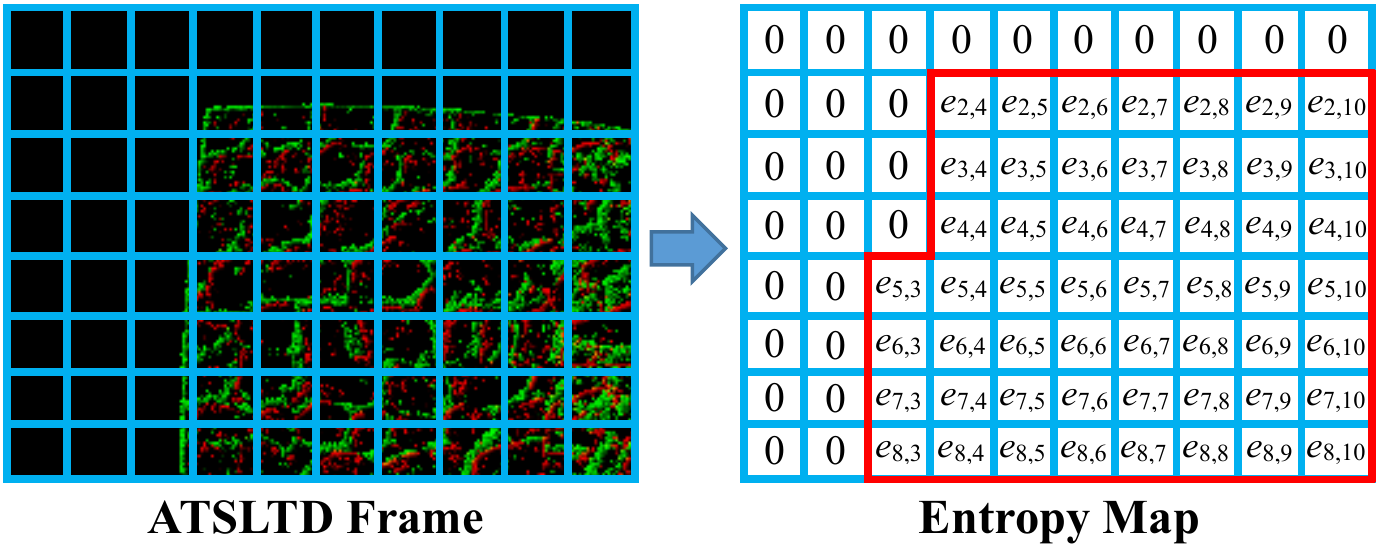}
	\end{center}
	\vspace{-3ex}
	\caption{An illustration of the non-zero grid entropy calculation. The left and right sub-figures show an ATSLTD frame and the corresponding entropy map. The grids with non-zero entropy values are in the red zone of the map.}
	\label{fig:nzge}
	\vspace{-4ex}
\end{figure}


\begin{figure*}
	\begin{center}
		\includegraphics[width=0.80\linewidth]{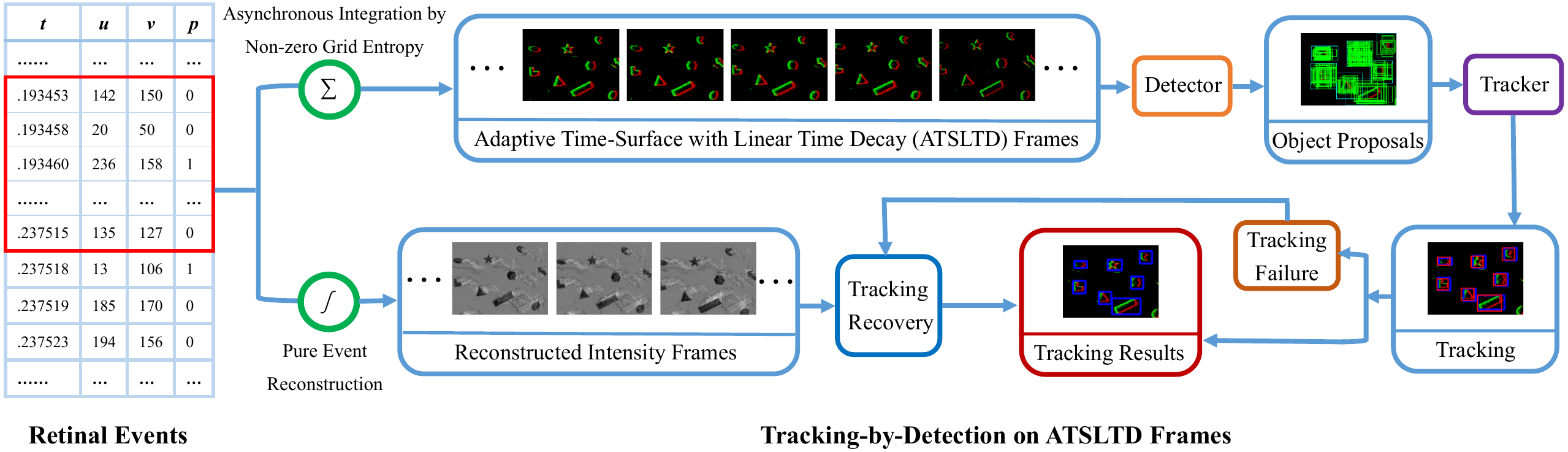}
	\end{center}
	\vspace{-3ex}
	\caption{The pipeline of the proposed ETD method. Initially, retinal events are asynchronously warped to a sequence of ATSLTD frames. Then the detector of ETD generates object proposals on each of the ATSLTD frames. And the tracker of ETD selects the best bounding box from the generated object proposals as the tracking result. Finally, if the tracker loses the tracked object, ETD will use the intensity frames, which are reconstructed from the retinal events, to recover tracking from the failure.}
	\label{fig:pipeline}
	\vspace{-3ex}
\end{figure*}

\subsection{Confidence Interval}
\label{subsec:ConfidenceInterval}
As discussed in Sec. \ref{subsec:ATSLTD}, effective event-based frame representations should have sharp and clear object contours for moving objects. Before calculating the confidence interval when the proposed asynchronous ATSLTD event-to-frame conversion algorithm is implemented, we need a metric to measure the displacement of the object contours in ATSLTD frames. Since the image entropy is a natural way to measure the amount of information in ATSLTD frames, in this work, we propose the Non-Zero Grid Entropy (NZGE) measure as the metric. To calculate the NZGE value of an ATSLTD frame, we divide the ATSLTD frame into $p \times q$ grids, each of which is a $r \times r$ image patch. Then we compute an image entropy for each of the grids to build an entropy map, as shown in Fig. \ref{fig:nzge}. Finally the NZGE value of the $i$-th ATSLTD frame is calculated as follow:
\begin{equation}
NZG{E_i} = \frac{1}{{n_i^{grid}}}\sum\limits_{x = 1}^p {\sum\limits_{y = 1}^q {entropy_{x,y}^i} },
\end{equation} 
where $n_i^{grid}$ is the number of the grids with non-zero entropy values. $entropy_{x,y}^i$ is the image entropy of the image patch in the $x$-th row and $y$-th column. The image entropy is defined as follow:
\begin{equation}
entropy_{x,y} =  - \sum\limits_{z = 0}^{255} {prob_{_z}^{x,y}{{\log }}\,} prob_{_z}^{x,y},
\end{equation}
where $prob_{_z}^{x,y}$ is the probability of a pixel having gray level $z$ in the $x$-th row and $y$-th column of the image patch (the range of the gray level are 0 to 255). The parameters $p$, $q$ and $r$, are respectively set to 45, 60 and 4 in this work. For the proposed two-channel ATSLTD frame representation, we respectively calculate two image entropy values for the two channels. The mean of the two image entropy values will be the image entropy of the two-channel ATSLTD frame. 

Event cameras will filter out the static part of a scene from their output. Thus, if we calculate an image entropy value over the whole ATSLTD frame, the zero entropy part of the ATSLTD frame will degrade the accuracy of the metric. Otherwise, if we calculate an image entropy value only for the pixels with non-zero gray levels, some small sensor noise may dominate the image entropy value. Therefore, we propose the NZGE to measure the amount of information in ATSLTD frames. The amount of information in an ATSLTD frame will increase along with object motions. As a result, we can use a confidence interval of NZGE values to maintain the object contour displacement in an ATSLTD frame within a reasonable range, which creates clear and sharp object contours for object detection and object tracking.

To calculate the lower and upper bounds of the confidence interval, we collect a set of NZGE values ${\mathcal C} = \{ {c_1},{c_2}, \ldots ,{c_{{n^{s}}}}\}$ from a set of sampled ATSLTD frames with a variety of clear and sharp object contours. The sample mean and the sample variance of ${\mathcal C}$ are $\overline {\mathcal C}$ and ${S}_{\mathcal C}^2$, respectively. We thereby can use the set of suitable NZGE values to create clear and sharp object contours. We assume that the suitable NZGE values are independent and distributed as a normal distribution $\mathcal N(\mu ,{\sigma ^2})$. Thus, ${\mathcal C}$ is ${n^{s}}$ observations from the normal distribution. Here we define a pivotal quantity $g$ as:
\begin{equation}
\label{eqn:tdis}
g = \frac{{\overline {\mathcal C} - \mu }}{{{S_{\mathcal C}}/\sqrt {n^{s}} }}
\end{equation} 
As a result, $g$ follows the $t$-distribution $t({n^{s}} - 1)$ with ${n^{s}}-1$ degrees of freedom. From Eq. (\ref{eqn:tdis}), we can estimate a confidence interval for the mean $\mu$ of the normal distribution $\mathcal N(\mu ,{\sigma ^2})$, as follow:
\begin{equation}
\overline {\mathcal C} - \left| {{g_{\omega /2}}} \right|\frac{{S_{\mathcal C}}}{{\sqrt {n^{s}} }} < \mu  < \overline {\mathcal C} + \left| {{g_{\omega /2}}} \right|\frac{{S_{\mathcal C}}}{{\sqrt {n^{s}} }},
\end{equation}
where $\omega$ is a two-sided significance level. We use $\omega = 0.05$ in this work, which means:
\begin{equation}
\Pr \left( {\overline {\mathcal C}  - \left| {{g_{0.025}}} \right|\frac{{{S_{\mathcal C}}}}{{\sqrt {{n^s}} }} < \mu  < \overline {\mathcal C}  + \left| {{g_{0.025}}} \right|\frac{{{S_{\mathcal C}}}}{{\sqrt {{n^s}} }}} \right) = 0.95
\end{equation}
Finally, the calculated confidence interval for NZGE values is:
\begin{equation} 
[\alpha, \beta] = \left[ {\overline {\mathcal C}  - \left| {{g_{0.025}}} \right|\frac{{{S_{\mathcal C}}}}{{\sqrt {{n^s}} }},\,\overline {\mathcal C}  + \left| {{g_{0.025}}} \right|\frac{{{S_{\mathcal C}}}}{{\sqrt {{n^s}} }}} \right]
\end{equation}
In this work, we collect ${n^{s}}=100$ samples with $\overline {\mathcal C} = 0.08795$ and ${S_{\mathcal C}} = 0.02394$ for the calculation of the confidence interval. According to the $t$-distribution table, ${g_{0.025}} = 1.984$. Therefore, the calculated confidence interval $[\alpha, \beta] \approx [0.0832, 0.0927]$. The ATSLTD frames, whose NZGE values lie in the calculated confidence interval, will be more likely to have clear and sharp object contours. Thus the calculated confidence interval is used in the ATSLTD event-to-frame conversion algorithm to trigger the ATSLTD frame generation effectively and asynchronously.

\subsection{Event-based Tracking-by-Detection}
\label{subsec:ETD}
Since the generated ATSLTD frames have shown clear object contours and they have a high spatio-temporal resolution, the proposed event-based tracking-by-detection method does not need complicated mechanisms, which may increase the computational cost. In this work, we propose an effective and efficient two-stage Event-based Tracking-by-Detection (ETD) method, as shown in Fig. \ref{fig:pipeline}. In the first stage, the detector of the proposed ETD generates a set of object proposals on each of the sequential ATSLTD frames for the target object specified in the first ATSLTD frame. Then the ETD method uses an Intersection over Union (IoU)-based tracker to select the best object proposal from the generated object proposals as the tracking result in the current frame.

\begin{table*}
	\caption{The details of the mixed event dataset. The FM, BC, SO, HDR and OC, in the Challenges column, are fast motion, background clutter, small object, HDR scene and occlusion, respectively.}
	\vspace{-3ex}
	\label{tab:dataset}
	\begin{center}
		\begin{tabular}{|l|l|c|c|}
			\hline
			Dataset & Sequence names & Feature & Challenges\\
			\hline\hline
			ECD     & shapes\_translation   & B\&W shape objects mainly with translations    & FM \\
			ECD     & shapes\_6dof          & B\&W shape objects with various 6-DoF motions  & FM \\
			ECD     & poster\_6dof          & Natural textures with cluttered background and various 6-DoF motions  & FM+BC \\
			ECD     & slider\_depth         & Various artifacts at different depths with only translations  & BC  \\ \hline
			EED     & fast\_drone           & A fast moving drone under a very low illumination condition  & FM+SO+HDR   \\
			EED     & light\_variations     & Same with the upper one with extra periodical abrupt flash lights  & FM+SO+HDR   \\
			EED     & what\_is\_background  & A thrown ball with a dense net as foreground  & FM+OC   \\
			EED     & occlusions            & A thrown ball with a short occlusion under a dark environment  & FM+OC+HDR    \\
			\hline
		\end{tabular}
	\end{center}
	\vspace{-4ex}
\end{table*}

The generated ATSLTD frames have recorded clear and sharp object contours. Thus object contour-based detectors can take advantage of the input ATSLTD frames. In the detection stage, we exploit EdgeBoxes \cite{zitnick2014edge} as the detector of the proposed ETD method for its high recall rate and fast speed, and it is used to generate object proposals. Each of the generated object proposals is an object bounding box in the current $i$-th ATSLTD frame $\mathcal{F}_{i}$ for the bounding box ${O_{i-1}}$ of the target object in the previous ($i-1$)-th ATSLTD frame $\mathcal{F}_{i-1}$. The center location and size of ${O_{i-1}}$ are ${c_{i-1}}$ and $(w_{i-1}, h_{i-1})$, respectively. We only detect object proposals within a searching region ${\mathcal R_{i}}$ for high computational speed. The center location and size of ${\mathcal R_{i}}$ are ${c_{i-1}}$ and $(\tau w_{i-1}, \tau h_{i-1})$. $\tau$ is a parameter to render the searching region slightly larger than the previous size.

By leveraging the high spatio-temporal resolution property of the ATSLTD frames, we can improve the generated object proposals to create a set of refined object proposals for tracking. For each of the object proposals, we compute a score between ${O_{i-1}}$ and the bounding box ${P_{i}}$ of the object proposal. The score between ${O_{i-1}}$ and ${P_{i}}$ is calculated as follows:
\begin{equation}
\label{eq:score} 
{\rm{score}}({O_{i-1}},{P_{i}}) = \phi \left( {\frac{{{w_{i-1}} \times {h_{i-1}}}}{{{w_{p_{i}}} \times {h_{p_{i}}}}}} \right) \times \phi \left( {\frac{{{w_{i-1}}/{h_{i-1}}}}{{{w_{p_{i}}}/{h_{p_{i}}}}}} \right),
\end{equation}
where ${w_{i-1}}$ and ${h_{i-1}}$ are the width and height of ${O_{i-1}}$, ${w_{p_{i}}}$ and ${h_{p_{i}}}$ are the width and height of ${P_{i}}$, and the function $\phi ( \cdot )$ is defined as:
\begin{equation} 
\phi (x) = \left\{ {\begin{array}{*{20}{c}}
	x\\
	{1/x}
	\end{array}} \right.{\rm{  }}\begin{array}{*{20}{c}}
{0 < x < 1}\\
{x \ge 1}
\end{array}
\end{equation}
Eq. (\ref{eq:score}) indicates that if ${P_{i}}$ is a refined object proposal of ${O_{i-1}}$, they should not have much change in terms of area and aspect ratio. After scoring all the generated object proposals, these object proposals are filtered to a set of refined object proposals that have a higher score than a threshold $\lambda$.

After the detection stage, we have a set of refined object proposals, which are the candidates of the target object in the current ATSLTD frame. The tracker of the proposed ETD method will further refine the candidates to find the best candidate for the target object. As described in Section \ref{subsec:ConfidenceInterval}, the object motion between every two adjacent ATSLTD frames is constrained to a moderate level in the ATSLTD event-to-frame conversion. Therefore the bounding boxes for the target object from two adjacent ATSLTD frames should have a large overlap with each other, which is suitable for IoU-based trackers (such as \cite{bochinski2017high}). The IoU measure, between the two bounding boxes ${O_{i-1}}$ and ${O_{i}}$, is defined as:
\begin{equation} 
{\rm{IoU}}({O_{i-1}},{O_{i}}) = \frac{{Area({O_{i-1}}) \cap Area({O_{i}})}}{{Area({O_{i-1}}) \cup Area({O_{i}})}}
\end{equation} 
In the tracking stage, we exploit a simple yet effective IoU-based tracker, which chooses the candidate bounding box that has the largest IoU with the previous bounding box ${O_{i-1}}$, as the estimated bounding box ${O_{i}}$ of the target object in the $i$-th ATSLTD frame $\mathcal{F}_{i}$.

For the tracking failure situation, the proposed ETD method also provides a tracking recovery strategy for it, as shown in Fig. \ref{fig:pipeline}. Along with the ATSLTD frames, there are sequential synchronized intensity event frames, which are reconstructed using the efficient pure event reconstruction method in \cite{scheerlinck2018continuous}. If an IoU between an estimated bounding box and its previous bounding box, is under a threshold $\mu$, we treat this situation as a tracking failure. And we use the DaSiamRPN tracker \cite{zhu2018distractor} to track the target object again on the reconstructed intensity event frames for recovering the tracking status from the tracking failure.


\section{Experiments}
\label{sec:Experiments}


\subsection{Experimental Settings}
\label{subsec:Pre-processing}


\begin{table*}
	\caption{Results obtained by the nine competing methods and the proposed ETD method on the four event sequences from the ECD dataset. The best values are highlighted by bold.}
	\vspace{-3ex}
	\label{tab:ecd_results}
	\begin{center}
		\begin{tabular}{|c|c|c|c|c|c|c|c|c|}
			\hline
			\multirow{2}{*}{Method} & \multicolumn{2}{c|}{shapes\_translation} & \multicolumn{2}{c|}{shapes\_6dof}   & \multicolumn{2}{c|}{poster\_6dof}   & \multicolumn{2}{c|}{slider\_depth} \\ \cline{2-9}
			& AP                 & AR                 & AP              & AR               & AP              & AR               & AP                & AR            \\ \hline\hline
			KCF\cite{henriques2015high}                     & 0.306             & 0.468            & 0.307          & 0.430          & 0.345          & 0.472          & 0.674            & 0.780  \\ \hline
			TLD\cite{kalal2012tracking}                     & 0.468             & 0.425            & 0.430          & 0.424          & 0.472          & 0.554          & 0.504            & 0.643       \\ \hline
			SiamFC\cite{bertinetto2016fully}                  & 0.685             & 0.872            & 0.668          & 0.842          & 0.585          & 0.726          & 0.806            & 0.910       \\ \hline
			ECO\cite{danelljan2017eco}                     & 0.746               & 0.931              & 0.717            & 0.877            & 0.614            & 0.719            & \textbf{0.915}              & 0.667         \\ \hline
			DaSiamRPN\cite{zhu2018distractor}                     & 0.668               & 0.878              & 0.642            & 0.831            & 0.517            & 0.651            & 0.713              & 0.890         \\ \hline
			DaSiamRPN-E\cite{zhu2018distractor}                     & 0.728               & 0.913              & 0.749            & 0.878            & 0.692            & 0.814            & 0.803              & 0.974         \\ \hline
			E-MS\cite{barranco2018real}                  & 0.675    & 0.768   & 0.612 & 0.668 & 0.417 & 0.373 & 0.447   & 0.350  \\ \hline
			ETD-FTW                  & 0.734    & 0.922   & 0.727 & 0.924 & 0.707 & 0.918 & 0.636   & 0.766  \\ \hline
			ETD-NR                  & 0.793    & 0.982   & 0.783 & 0.972 & 0.775 & 0.992 & 0.803   & 0.983  \\ \hline
			ETD                  & \textbf{0.817}    & \textbf{0.998}   & \textbf{0.809} & \textbf{0.998} & \textbf{0.788} & \textbf{0.995} & 0.816   & \textbf{0.997}  \\ \hline
		\end{tabular}
	\end{center}
	\vspace{-2ex}
\end{table*}


\begin{table*}
	\caption{Results obtained by the nine competing methods and the proposed ETD method on the four event sequences of the EED dataset.  The best values are highlighted by bold.}
	\vspace{-3ex}
	\label{tab:eed_results}
	\begin{center}
		\begin{tabular}{|c|c|c|c|c|c|c|c|c|}
			\hline
			\multirow{2}{*}{Method} & \multicolumn{2}{c|}{fast\_drone} & \multicolumn{2}{c|}{light\_variations} & \multicolumn{2}{c|}{what\_is\_background} & \multicolumn{2}{c|}{occlusions} \\ \cline{2-9}
			& AP               & AR           & AP                & AR                & AP                   & AR                & AP              & AR           \\ \hline\hline
			KCF\cite{henriques2015high}                     & 0.169           & 0.176      & 0.107            & 0.066           & 0.028               & 0.000               & 0.004          & 0.000          \\ \hline
			TLD\cite{kalal2012tracking}                     & 0.315           & 0.118      & 0.045            & 0.066           & 0.269      & 0.333      & 0.092          & 0.167      \\ \hline
			SiamFC\cite{bertinetto2016fully}                  & 0.559           & 0.667      & 0.599            & 0.675           & 0.307               & 0.308           & 0.148          & 0.000          \\ \hline
			ECO\cite{danelljan2017eco}                     & 0.637           & 0.833      & 0.586            & 0.688           & 0.616               & 0.692           & 0.108          & 0.143          \\ \hline
			DaSiamRPN\cite{zhu2018distractor}                     & 0.673           & 0.853      & 0.654            & 0.894           & \textbf{0.678}               & \textbf{0.833}           & 0.189          & 0.333          \\ \hline
			DaSiamRPN-E\cite{zhu2018distractor}                      & 0.646           & 0.847      & 0.705            & 0.902           & 0.573               & 0.742           & 0.373          & 0.605          \\ \hline
			E-MS\cite{barranco2018real}                  & 0.313  & 0.307 & 0.325   & 0.321  & 0.362               & 0.360           & 0.356 & 0.353 \\ \hline
			ETD-FTW                  & 0.576    & 0.673   & 0.722 & 0.874 & 0.562 & 0.733 & 0.263   & 0.533  \\ \hline
			ETD-NR                  & 0.722    & 0.883   & 0.833 & 0.925 & 0.638 & 0.781 & 0.414   & 0.622  \\ \hline
			ETD                  & \textbf{0.738}  & \textbf{0.897} & \textbf{0.842}   & \textbf{0.933}  & 0.653               & 0.807           & \textbf{0.431} & \textbf{0.647} \\ \hline
		\end{tabular}
	\end{center}
	\vspace{-3ex}
\end{table*}


In this section, we evaluate the proposed ETD method and nine competing methods on a challenging mixed event dataset, which includes a part of the Event Camera Dataset (ECD) \cite{mueggler2017event} and the Extreme Event Dataset (EED) \cite{mitrokhin2018event}. The two datasets were recorded using a DAVIS \cite{brandli2014240} event camera in real-world environments. The details of the mixed event dataset are shown in Table \ref{tab:dataset}. Note that the mixed event dataset consists of both the event data sequences and the corresponding video sequences for each sequence. Since the ECD dataset does not provide the ground truth for object tracking, we manually label a rectangle bounding box for each object as the ground truth in the ECD dataset.

In this work, we select nine competing object tracking methods, including KCF \cite{henriques2015high}, TLD \cite{kalal2012tracking}, SiamFC \cite{bertinetto2016fully}, ECO \cite{danelljan2017eco}, DaSiamRPN \cite{zhu2018distractor}, an event-based variant of DaSiamRPN \cite{zhu2018distractor} (called as DaSiamRPN-E), E-MS \cite{barranco2018real} and two variants of our ETD method (called as ETD-FTW and ETD-NR) as the competitors. About these competitors: KCF \cite{henriques2015high} is an effective correlation filter-based object tracking method. TLD \cite{kalal2012tracking} is a classical and robust Tracking-by-Detection method. SiamFC \cite{bertinetto2016fully} is an efficient and effective object tracking method, which uses a fully-convolutional Siamese network to track objects. ECO \cite{danelljan2017eco} is a state-of-the-art object tracking method that employs compact samples to train continuous convolutional operators for visual tracking. DaSiamRPN \cite{zhu2018distractor} is a state-of-the-art tracking method that improves the fully-convolutional siamese network by using distractor-aware training. DaSiamRPN-E, which uses ATSLTD frames as its input, is an event-based variant of DaSiamRPN. E-MS \cite{barranco2018real} is a state-of-the-art event-based target tracking method based on both mean-shift clustering and Kalman tracking. Here the E-MS method is extended to support bounding box-based evaluation by using the minimum enclosing bounding boxes of those events belonging to the same cluster center as its tracking results. ETD-FTW and ETD-NR are two variants of ETD. ETD-FTW uses the Fixed Time Window (FTW) frames (as shown in Fig. \ref{fig:conversion}) instead of using the ATSLTD frames as its input. ETD-NR removes the tracking recovery part from the original ETD method. The two variants are used to show the influence of the components of the proposed ETD method on its performance.

For KCF and TLD, we use their OpenCV implementations with the suggested settings. For SiamFC and ECO, we use their original Matlab implementations with the default parameters. For DaSiamRPN and E-MS, we respectively use their original Python and C++ implementations with the default parameters. For the proposed ETD method, the maxBoxes and minBoxArea of the Edgeboxes detector are respectively set to 1000 and 100 considering the balance between speed and accuracy. The parameter $\tau$ for the searching region is set to 1.5. The threshold $\lambda$ for refining object proposals is set to 0.7. The threshold $\mu$, which is used to judge the tracking failure situation, is set to 0.3. We fix these parameter values for all the following experiments.

\subsection{Evaluation Metrics}
\label{subsec:EvaluationMetrics}
During the evaluation, we run all of the competing methods ${N^{rep}}$ times. If a sequence of the dataset includes multiple objects, we evaluate all the objects separately. If a tracking failure occurs for a competing method, we will reinitialize the method at the next frame. To evaluate the precision of all the methods, we calculate the Average Precision (AP) as follow:
\begin{equation}
\label{eq:aor}
{AP} = \frac{1}{{{N^{rep}}}}\frac{1}{{{N^{obj}}}}\sum\limits_{a = 1}^{{N^{rep}}} {\sum\limits_{b = 1}^{{N^{obj}}} {\frac{{O_{a,b}^E \cap O_{a,b}^G}}{{O_{a,b}^E \cup O_{a,b}^G}}}},
\end{equation}
where $O_{a,b}^{E}$ is the estimated bounding box in the $a$-th round of the evaluation for the $b$-th object, and $O_{a,b}^{G}$ is the corresponding ground truth. ${N^{rep}}$ is the repeat times of the evaluation, and ${N^{obj}}$ is the number of objects in the current sequence. We set ${N^{rep}}$ to 5.

\begin{figure*}
	\begin{center}
		\includegraphics[width=0.80\linewidth]{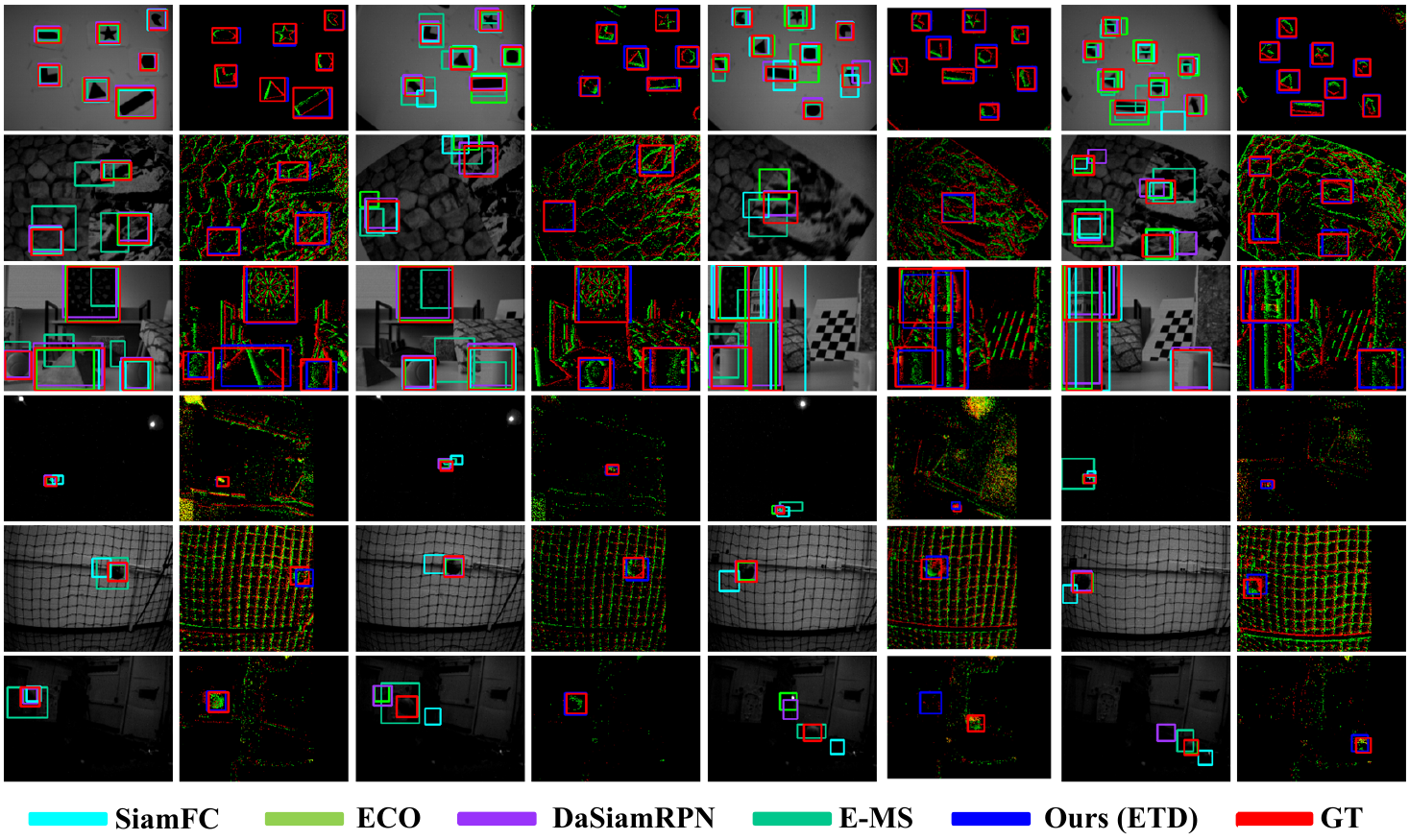}
	\end{center}
	\vspace{-3ex}
	\caption{Tracking results obtained by SiamFC, ECO, DasiamRPN, E-MS and the proposed ETD method. Each row shows a representative sequence of the mixed event dataset. From top to bottom, the sequences are \emph{shape\_6dof}, \emph{poster\_6dof}, \emph{slider\_depth}, \emph{light\_variations}, \emph{what\_is\_background} and \emph{occlusions}, respectively. From left to right, the first, third, fifth and seventh columns respectively show the results of SiamFC, ECO, DaSiamRPN and E-MS. The second, fourth, sixth and eighth columns respectively show the results of the proposed ETD method. Best viewed in color.}
	\label{fig:results}
	\vspace{-4ex}
\end{figure*}


We also calculate the Average Robustness (AR) to measure the robustness of all the methods as follow:
\begin{equation}
{AR} = \frac{1}{{{N^{rep}}}}\frac{1}{{{N^{obj}}}}\sum\limits_{a = 1}^{{N^{rep}}} {\sum\limits_{b = 1}^{{N^{obj}}} {succes{s_{a,b}}}},
\end{equation}
where $succes{s_{a,b}}$ indicates that whether the tracking in the $a$-th round for the $b$-th object is successful or not (1 means success and 0 means failure). If the AP value obtained by a method for an object is under 0.5, we will consider it as a tracking failure case.


\subsection{Evaluation on the Mixed Event Dataset}
In the mixed event dataset, we use the \emph{shapes\_translation}, \emph{shapes\_6dof}, \emph{poster\_6dof} and \emph{slider\_depth} sequences as the representative sequences of the ECD dataset \cite{mueggler2017event} for comparison. The first three sequences include fast object motions. The third and fourth sequences include complicated background clutters. And the object textures in the four sequences vary from simple B\&W shapes to complicated artifacts. For these sequences, we mainly concern about the performance of all methods for a variety of object motions, especially for fast motion, and for background clutter.

The quantitative results obtained by the competing methods are given in Table \ref{tab:ecd_results}. Moreover, some representative qualitative results obtained by SiamFC, ECO, DaSiamRPN, E-MS and the proposed ETD are shown in the top three rows of Fig. \ref{fig:results}. From Table \ref{tab:ecd_results}, we can see that the proposed ETD achieves the best performance on the first three sequences and it achieves the best AR and the second best AP on the fourth sequence. In comparison, ECO achieves the best AP on the fourth sequence. However, it has a relatively low AR compared with the proposed ETD. We find that it is the reinitialization protocol that helps ECO to achieve the high AP. In addition, as shown in Fig. \ref{fig:results}, the proposed ETD has achieved better performance in handling fast object motions. In comparison, KCF, TLD, SiamFC, ECO and DaSiamRPN usually achieve low precision values, due to the influence of motion blur. E-MS can handle most fast motions. However, it is less effective to handle cluttered backgrounds (e.g., for the \emph{poster\_6dof} and \emph{slider\_depth} sequences). As a variant of DaSiamRPN, DaSiamRPN-E achieves satisfying results by exploiting the virtues of the ATSLTD representation. Comparing with the proposed ETD method, ETD-FTW and ETD-NR have achieved inferior performance, which shows that both the ATSLTD representation and the tracking recovery component can improve the performance of the proposed ETD. However, as the quantitative results show, the ATSLTD representation has more influence on the performance of ETD than the tracking recovery component. In summary, the evaluation on the ECD dataset demonstrates the superiority of the proposed ETD in handling fast motion, various textured objects and cluttered backgrounds.

Moreover, we also use the recently proposed EED dataset to evaluate the competing methods. The EED dataset \cite{mitrokhin2018event} contains four sequences: \emph{fast\_drone}, \emph{light\_variations}, \emph{what\_is\_background} and \emph{occlusions}. The first three sequences respectively record a small and fast moving drone under HDR environments. The fourth sequence records a moving ball with a net as foreground. Using this dataset, we evaluate the influence of HDR scenes and occlusions on the performance of the competing methods.

The quantitative results are shown in Table \ref{tab:eed_results} and some representative qualitative results are shown in the bottom three rows of Fig. \ref{fig:results}. From the results, we can see that the proposed ETD achieves the best performance on most of the sequences except for the \emph{what\_is\_background} sequence, on which it obtains the second best AP and AR. This is because that the foreground net in the \emph{what\_is\_background} sequence partially occludes the tracked ball, which results in a negative influence on the corresponding contour of the tracked ball. Among the nine competitors, KCF, TLD, SiamFC, ECO and DaSiamRPN cannot effectively handle with fast motion and low illumination conditions. We also find that the performance of E-MS is significantly affected by the sensor noises in HDR environments. As a result, E-MS achieves poor results on the EED dataset. Moreover, ETD-FTW and ETD-NR are inferior to the proposed ETD, which shows the contribution of the ATSLTD representation and the tracking recovery component of the proposed ETD. Overall, the evaluation on the EED dataset shows that the proposed ETD can effectively handle HDR scenarios. 


\subsection{Time Cost}
\label{subsec:timecost}
The proposed method is implemented using Python on a PC with an Intel i7, 32G RAM and an NVIDIA GTX 1080 GPU. For the mixed event dataset, the average computational time for tracking an object is 21.97 ms per ATSLTD frame. Note that only the tracking recovery part of the proposed ETD method is accelerated by GPU, while the other parts of the ETD method work on CPU.

\section{Conclusion}
\label{sec:Conclusion}
In this paper, we present a novel event-to-frame conversion algorithm that can asynchronously warp the spatio-temporal information in sequential retinal events to ATSLTD frames. The event-to-frame conversion is driven by motions and it can record object contours more clearly on the generated ATSLTD frames, which facilitates to detect and track moving objects. Then, we propose an event-based tracking-by-detection method. It can effectively and efficiently track objects on the ATSLTD frames. Extensive experiments demonstrate the great advantages of the proposed ETD method for object tracking in challenging situations such as fast motion and HDR scenes.


%
\begin{acks}
This work is supported by the National Natural Science Foundation of China (Grant No. U1605252, 61872307, 61432014, 61772402 and 61671339).
\end{acks}

%
\bibliographystyle{ACM-Reference-Format}
\balance
\bibliography{egbib}

\end{document}